\documentclass[letterpaper, 10 pt, conference]{ieeeconf}  

\IEEEoverridecommandlockouts                              

\usepackage[T1]{fontenc}

\usepackage{amsmath} 
\usepackage{amssymb}  
\usepackage{booktabs, makecell, tabularx}       
\usepackage{graphicx}

\usepackage{multirow}

\usepackage{makecell}

\newcommand{\matr}[1]{\mathbf{#1}}

\usepackage{here}

\usepackage{lipsum}
\usepackage{xcolor}

\title{\LARGE \bf
Multimodal Anomaly Detection based on Deep Auto-Encoder \\for Object Slip Perception of Mobile Manipulation Robots
}

\author{Youngjae Yoo$^{1,2\dagger}$,
        Chung-Yeon Lee$^{3,4\dagger}$,
        and Byoung-Tak Zhang$^{1,2,3}$ 
\thanks{This work was supported by the Air Force Office of Scientific Research (FA2386-17-1-4128, 25\%), and partly supported by ICT R\&D program of MSIP/IITP (2015-0-00310(15\%), 2017-0-00162(15\%), 2017-0-01772(15\%), 2018-0-00622(15\%)) and Samsung Electronics (15\%). The authors would also like to thank Toyota Motor Corporation for supporting HSR gratefully.}%
\thanks{$^{1}$ Artificial Intelligence Institute, Seoul National University, Seoul, Korea}%
\thanks{$^{2}$ Interdisciplinary Program in Neuroscience, Seoul National University}%
\thanks{$^{3}$ Dept. of Computer Science and Engineering, Seoul National University}%
\thanks{$^{4}$ Surromind Inc., Seoul, Korea}%
\thanks{\tt\small \{yjyoo, cylee, btzhang\}@bi.snu.ac.kr}%
\thanks{$\dagger$ These two authors contribute equally to this work.}
}

\begin{document}

\maketitle
\thispagestyle{empty}
\pagestyle{empty}

%
%

\begin{abstract}

Object slip perception is essential for mobile manipulation robots to perform manipulation tasks reliably in the dynamic real-world.
Traditional approaches to robot arms' slip perception use tactile or vision sensors.
However, mobile robots still have to deal with noise in their sensor signals caused by the robot's movement in a changing environment.
To solve this problem, we present an anomaly detection method that utilizes multisensory data based on a deep autoencoder model.
The proposed framework integrates heterogeneous data streams collected from various robot sensors, including RGB and depth cameras, a microphone, and a force-torque sensor.
The integrated data is used to train a deep autoencoder to construct latent representations of the multisensory data that indicate the normal status.
Anomalies can then be identified by error scores measured by the difference between the trained encoder's latent values and the latent values of reconstructed input data.
In order to evaluate the proposed framework, we conducted an experiment that mimics an object slip by a mobile service robot operating in a real-world environment with diverse household objects and different moving patterns.
The experimental results verified that the proposed framework reliably detects anomalies in object slip situations despite various object types and robot behaviors, and visual and auditory noise in the environment.

\end{abstract}

%
%
\section{INTRODUCTION}

Recent developments in robotics and artificial intelligence technology have led to the creation of robots that substitute humans in simple service tasks.
These service robots have advanced rapidly from providing information at kiosks, to autonomous mobile agents working in several areas such as restaurants, airports, and museums.
It is anticipated that service robots will soon inevitably evolve to use robotic arms in order to perform more complex mobile manipulation tasks.

One essential skill to be considered before making practical use of robotic arms is slip perception: recognizing if a grasped object has slipped off the gripper or not \cite{howe1989sensing, james2018slip}.
Object slip may occur due to incomplete gripping, rapid movement of the robot, or collision with obstacles in the environment.
If the robot is unable to notice the slip, it will continue to move to finish the task without the object, thereby failing the task.
Many prior researches have investigated slip perception by using tactile sensors or cameras on a typical industrial robot fixed to a worktable \cite{zapata2019tactile, li2018slip}.
Despite making meaningful progress on this task, this avenue of research is impractical for mobile robots in real service fields for two reasons.
Firstly, not all robots with robotic arms have tactile sensors on their grippers.
Secondly, analyzing the image stream from a real environment is difficult due to the continuously changing background and various types of objects that the robot has to deal with.

There are many other factors that can indicate an object slip, such as the sound of the object hitting the ground and the loss of weight in the grasping hand. 
The goal of this study is to implement a slip perception model that utilizes the multimodal data collected by a mobile manipulation robot in a dynamic environment.

\begin{figure}[b]
    \centering
    \includegraphics[width=\columnwidth]{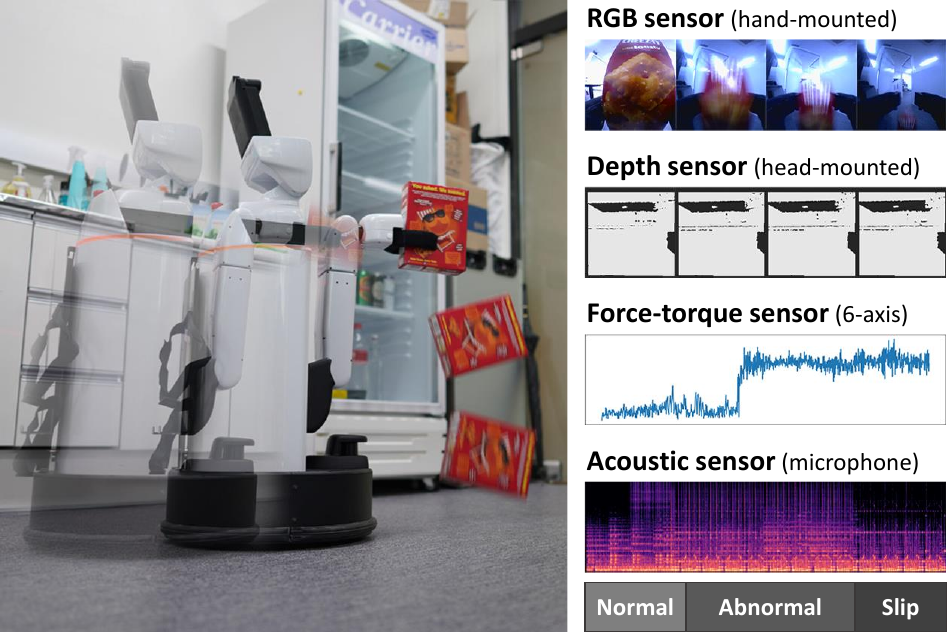}
    \caption{Example of an object slip perception experiment}
    \label{fig:scenario}
\end{figure}

In this paper, we propose a model that first provides multimodal integration by synchronizing and combining heterogeneous data from independent sensors of a robot. 
Then, the integrated data is fed into an autoencoder-based multimodal anomaly detection network, that can robustly detect a slip even in dynamic environmental changes.

To validate the proposed model, we designed an experiment in which a mobile robot collects multimodal sensor data while holding and losing grasp of an object as seen in Fig. \ref{fig:scenario}.
Everyday objects frequently used in our daily lives were tested in this experiment,
and visual and auditory noise was deliberately added to the experimental setting to mimic a real-life environment.
In addition, we present how the performance of the anomaly detection for each sensor data varies depending on the noises from the dynamic environment through ablation studies.

\begin{figure*}[t!]
    \centering
    \includegraphics[width=\textwidth]{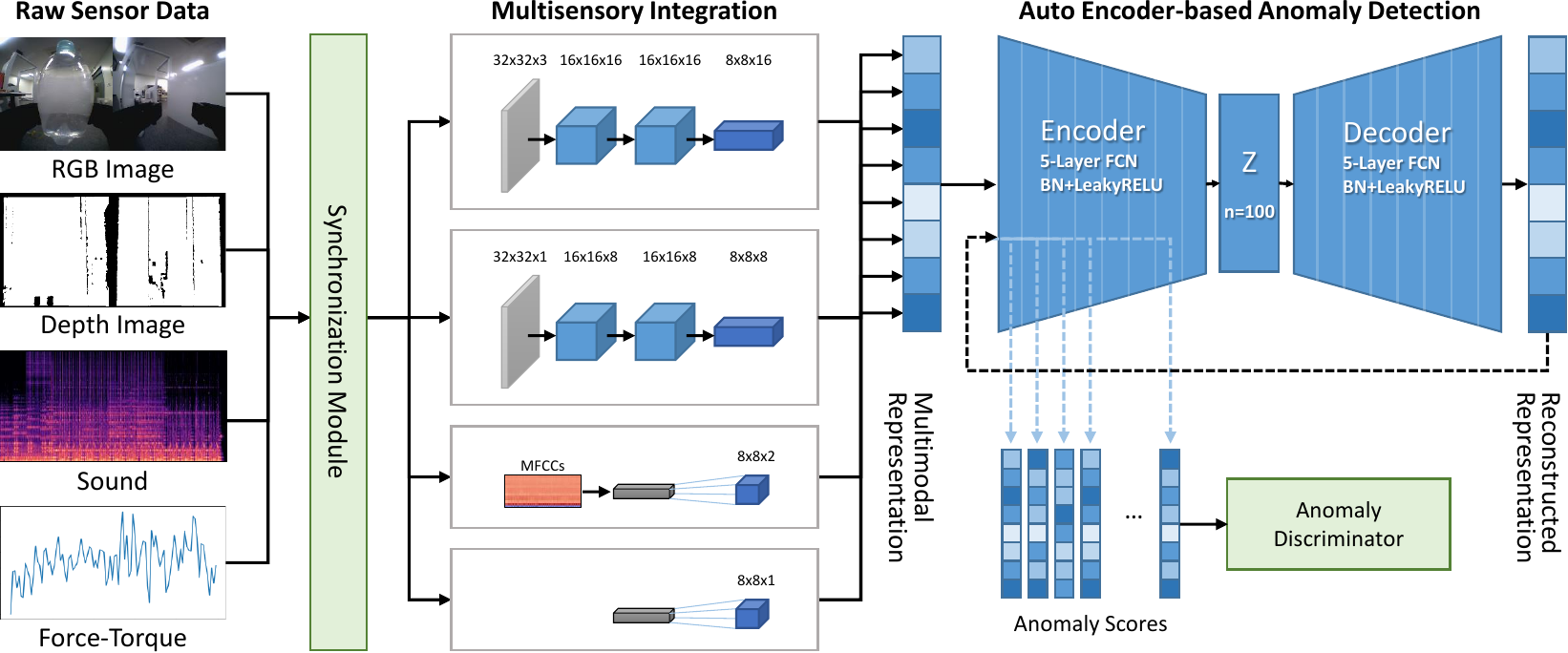}  
    \vspace{-.2cm}
    \caption{The overall architecture of the multimodal anomaly detection framework}
    \label{fig:framework}
\end{figure*}

%
%
\section{Related works}

\subsection{Grip Stability and Slip Prediction}
Improper orientation of a robot hand or insufficient contact pressure can cause gripped objects to slip. If a robot fails to detect the slip, it will continue its task without the object in its hand, inevitably failing to complete its intended task.
Many attempts have been made to solve this particular problem using tactile sensors, dating back to 1989 \cite{howe1989sensing} until today \cite{james2018slip}.
Recent developments in deep learning has brought with it attempts to exploit deep neural networks to address the slip problem \cite{zapata2019tactile, li2018slip}.
Li et al. \cite{li2018slip} demonstrated that tactile and visual information are complementary to each other in the slip detection task. Especially for objects with slippery and smooth surfaces, vision provided more cues to compensate for tactile sensors with poor detection accuracy.

\subsection{Anomaly Detection}
One way to evaluate grip success and stability is to apply anomaly detection on the robot's sensor data.
Anomaly detection has been a topic of interest in diverse research areas for a long time, and various methods have been proposed across multiple fields \cite{chandola2009anomaly, lee2000information, patcha2007overview, khalastchi2015online}.
Many recent studies focus on an autoencoder-based method that utilizes the reconstructed result's error as an anomaly score \cite{an2015variational, Kim2020RaPP, vasilev2018q}.
This method determines anomalies by comparing the difference between the input data and the data reconstructed by a trained autoencoder.
Kim et al. \cite{Kim2020RaPP} proposed a method that compares the hidden layer values of the encoder and decoder, instead of comparing just the input and output values of the autoencoder.
This approach has shown good performance in novelty detection on datasets in diverse domains including robotics malfunctions, steel surface defects, and hand written digits.

\subsection{Multimodal Approaches in Robot Manipulations}
Sensor signals on mobile robots in dynamic environments are extremely vulnerable to unwanted noise.
For example, robot acceleration can simultaneously affect weight sensors and cause irregularities in camera input due to motion blur and a rapidly changing background.
Additionally, dynamic environments will cause robot microphones to catch all kinds of sounds, which are impossible to expect in advance.

There have been attempts to approach problems caused by dynamic environments by integrating multisensory data so that each sensor data can complement each other's noise \cite{noda2014multimodal}.
Lee et al. \cite{lee2019making} proposed a self-supervised reinforcement learning model that encodes both tactile and visual sensors to operate a robotic arm in an unstructured environment.
Using force and sound data, Park et al. \cite{park2016multimodal, park2017multimodal, park2018multimodal} conducted multiple studies on anomaly detection in robot manipulations, such as pushing objects or feeding a person.

%
%
\section{Multimodal Anomaly Detection}

\subsection{Multimodal Data Integration}
Raw sensory inputs are heterogeneous data with different sizes and properties.
It is necessary to synchronize, normalize and integrate the multimodal data to feed them to the autoencoder network.
The synchronization module transforms all sensor data to 10Hz and synchronizes them to the nearest timestamp values.
Then, RGB and depth images are normalized to values ranging from 0 to 1 based on values between 0 and 255.
The acoustic and force-torque sensor data are normalized to values ranging from 0 to 1 based on the minimum and maximum values of each data type.
For auditory inputs, we use the mel-frequency cepstrum coefficients (MFCC \cite{davis1980comparison}) that were extracted at 0.1 second intervals from raw audio signals.
The synchronized data is then compressed, aligned, and combined through convolutional computation in multisensory integration as shown in Fig. \ref{fig:framework}.
Note that the multisensory integration in our framework is initialized by default settings of PyTorch \cite{pytorch}, and is not further optimized during the learning phase. 

\subsection{Autoencoder-based Anomaly Detection}
An autoencoder is a type of artificial neural network used to efficiently learn data encodings in an unsupervised manner \cite{Rumelhart1986, kramer1991nonlinear}.
An autoencoder can learn to encode the representation of the input data by training the neural network to ignore signal noise and only extract significant features.
The autoencoder can then use this reduced encoding to reconstruct an output close to the original input. 

\begin{figure}[t]
      \centering
      \includegraphics[width=1.0\columnwidth]{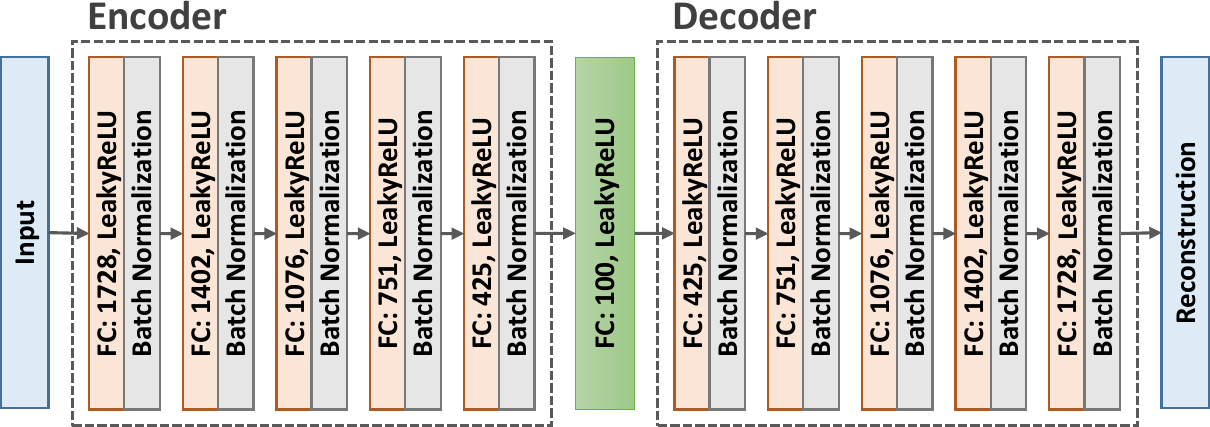}
      \caption{An autoencoder model used in the proposed framework}
      \label{fig:autoencoder}
\end{figure}

We use a symmetric architecture with fully-connected layers to build the proposed autoencoder model as shown in Fig. \ref{fig:autoencoder}.
Each encoder and decoder has 5 layers respectively, and we set the bottleneck size to 100.
Leaky-ReLU \cite{xu2015empirical} activation and batch normalization \cite{ioffe2015batch} layers are appended to all layers except the last layer for regularization.

We utilize the autoencoder to detect anomalies in sensory inputs.
An autoencoder trained with normal data cannot effectively compress and restore abnormal data.
Thus, the reconstruction error of the autoencoder can be used as an indicator to detect whether the input data is normal or abnormal.
The reconstruction error from the latent values of the autoencoder is derived as follows.

Let $H_l(x)$ be the value of the $l$-th hidden layer when the input $x$ passes through the encoder $f$, and $\hat{x}$ the output of the decoder $g$. Let $\hat{H}_l(x)$ be the $l$-th hidden layer after $\hat{x}$ passes through the encoder. $\hat{H}_l(x)$ can be expressed as:

\vspace{-.2cm}
\begin{equation}
    \label{eq:eq1}
    \begin{aligned}
        H_l(x) &= f_{1:l}(x),\\
        \hat{H}_l(x) &= f_{1:l}(\hat{x}) = f_{1:l}(g(f(x))).
    \end{aligned}
\end{equation}

\subsection{Anomaly Scores}
To calculate the anomaly score of the input data from the reconstruction error of the autoencoder, we use Normalized Aggregation along Pathway (NAP) scores \cite{Kim2020RaPP}.
NAP uses Singular Value Decomposition (SVD) \cite{golub1971singular} to transform the hidden-layer reconstruction error distribution into a normal distribution, and uses the absolute value of the error as the anomaly score.

Let $H(x)$ and $\hat{H}(x)$ be the concatenation of $H_1(x),\dots, H_L(x)$ and $\hat{H}_1(x),\dots, \hat{H}_L(x)$ respectively, and let $d(x) = H(x)-\hat{H}(x)$.
Let $\matr{D}$ be a matrix made up of the $d(s)$ values as row vectors for all $s\in S$ where $S$ refers to the training dataset, $\bar{\matr{D}}$ the column-wise centered matrix of $\matr{D}$, and $\mu$ the column-wise average of $\matr{D}$.
Then an NAP scores are calculated by:

\vspace{-.2cm}
\begin{equation}
    \label{eq:eq2}
    Score(x) = ||(d(x) - \mu)^T \matr{V}\matr{\Sigma}^{-1}||^2_2,
\end{equation}
where $\matr{V}$ and $\matr{\Sigma}$ are values from the SVD of $\bar{\matr{D}}$.

From the anomaly scores, we can determine how much the input data differs from the normal data, but the measure of the difference may vary depending on the task.
Thus it is important to find an optimal threshold that classifies anomalies for practical use of the anomaly score.
One approach to address this is the use of the receiver operating characteristics (ROC) curve of NAP to further interpret the predicted probabilities by comparing the distribution between the scores of normal and abnormal data.

Here, we use the area under the ROC (AUROC) to evaluate the performance of anomaly detectors.
Since abnormal data have an inherent characteristic that is fairly imbalanced with respect to normal data, we also use the area under the precision-recall curve (AUPRC) and F1-score to see robustness of detecting anomalies in such imbalanced data.

\section{Experimental Setup}

\subsection{Robot Platform}
\begin{figure}[!htb]
    \centering
    \includegraphics[width=0.8\columnwidth]{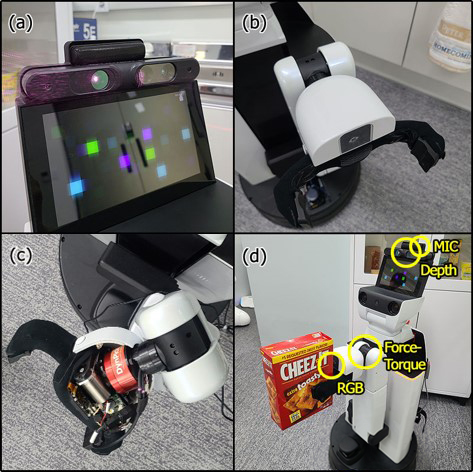}
    \caption{Type and location of sensors on the HSR robot}
    \label{fig:sensors}
\end{figure}

The robot we used for the experiment is the Human Support Robot (HSR \cite{hsr}), a mobile robot manufactured by Toyota depicted in Fig. \ref{fig:sensors}.
We developed a program that allows the robot to grasp, move, drop an object, and record sensor data in the process based on ROS.
The height of the HSR is 100-135cm (tallest when lifting the upper body), and is equipped with a gripper on its arm for manipulation tasks.
The equipped differential wheels and a central yaw-joint allows the robot to move omnidirectionally up to 20$\,$cm/s.
The sensors used in the experiment are shown in Fig. \ref{fig:sensors}: (a) the head 3D sensor (ASUS Xtion) and the microphone, (b) the hand RGB camera and (c) the 6-axis force-torque sensor (DynPick WDF-6M200-3).
The installed location of each sensor is indicated in Fig. \ref{fig:sensors}(d).

\subsection{Objects for Slip Detection Task}
The experiment was designed to imitate an object slip by a mobile service robot operating in a real-life environment.
As such, we carefully selected everyday objects that were frequently used in daily lives.
Additionally, the target objects were chosen so that they vary in characteristics, such as size, shape, weight, transparency, texture, and rigidity.
This was to prevent the model from learning biased features caused by specific objects.
Objects that are too large to fit in the gripper (13.5cm in width) of the HSR were excluded.
In order to make the experiment scalable, the objects used in the experiment all either matched or were similar to the objects from the YCB dataset \cite{calli2015ycb}.
The list of the type and weight for the selected objects is shown in Table \ref{tab:datasets}.

\begin{table}[t]
    \centering
    \caption{List of objects in our dataset for the object slip task}
    \label{tab:datasets}
    \begin{tabularx}{\linewidth}{@{}lcrl@{}}
        \toprule
        \thead[lb]{Object} & \thead[lb]{Photo} & \thead[lb]{Weight} & \thead[lb]{Property}\\
        \midrule
    Cracker box &
        \includegraphics[width=0.45in, height=0.3in]{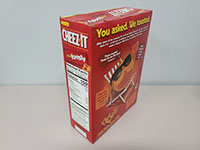}
        & 421g & Large size and heavy weight \\
    Bag of cookies &
        \includegraphics[width=0.45in, height=0.3in]{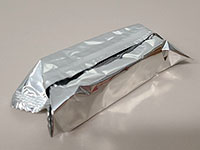}
        & 30g & Light reflectivity \\
    Furry toy &
        \includegraphics[width=0.45in, height=0.3in]{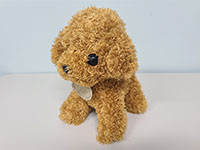}
        & 102g & High variability, low rigidity\\
    Book &
        \includegraphics[width=0.45in, height=0.3in]{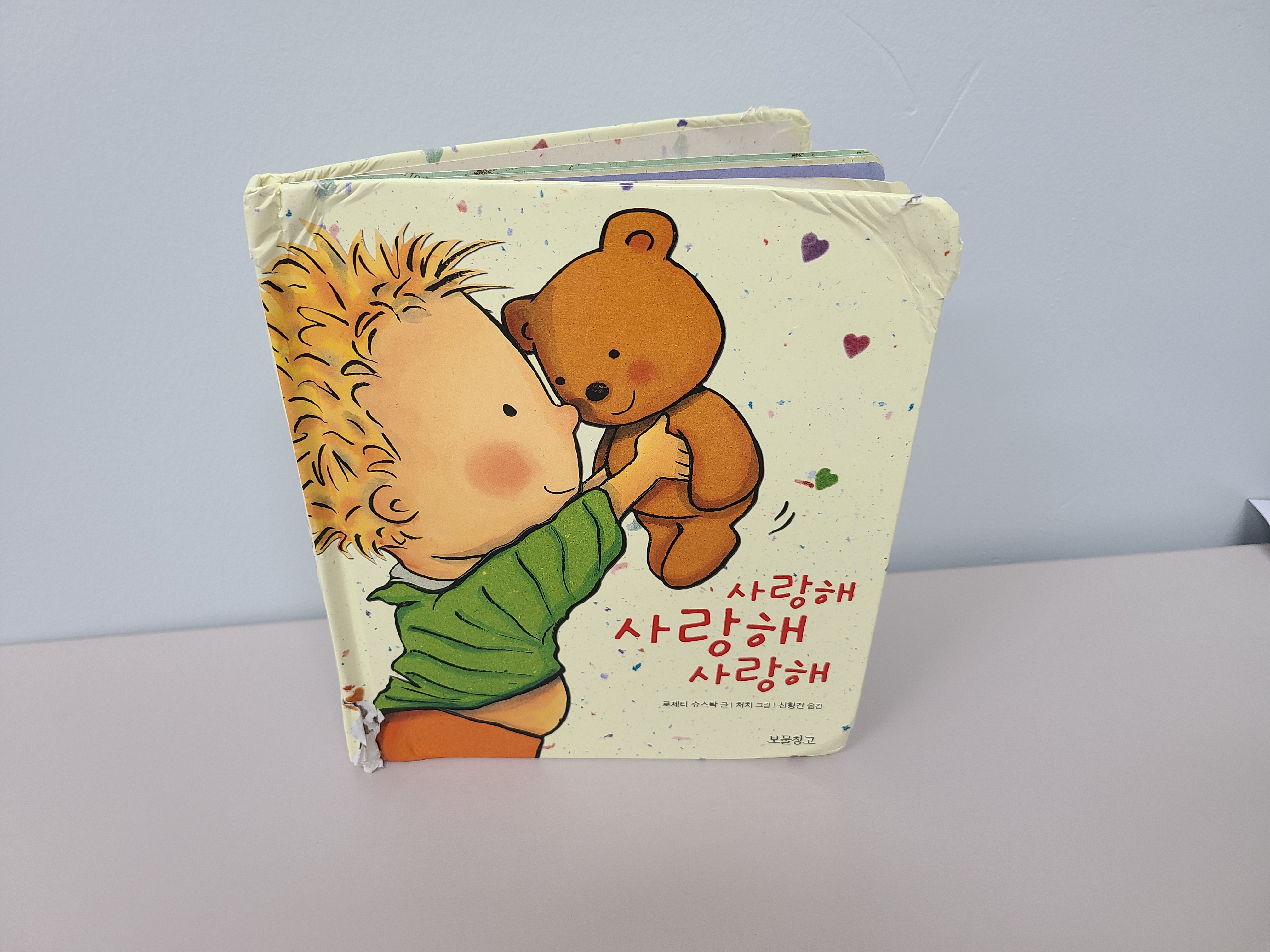}
        & 214g & Thin and hard covered \\
    Metal cup &
        \includegraphics[width=0.45in, height=0.3in]{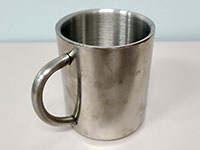}
        & 118g & High rigidity\\
    Plastic plate &
        \includegraphics[width=0.45in, height=0.3in]{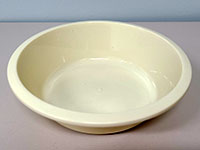}
        & 38g & Symmetry, elasticity \\
    Board eraser &
        \includegraphics[width=0.45in, height=0.3in]{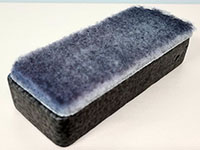}
        & 10g & Light weight\\
    Plastic bottle &
        \includegraphics[width=0.45in, height=0.3in]{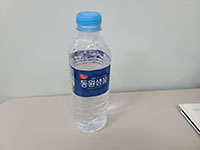}
        & 423g & High transparency\\
  \bottomrule
\end{tabularx}
\end{table}

\subsection{Experimental Protocols}

\begin{figure}[t]
    \centering
    \includegraphics[width=1.0\columnwidth]{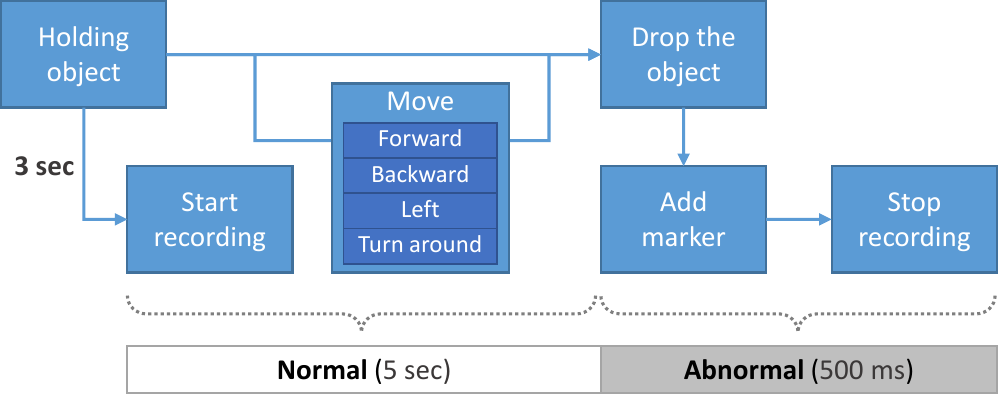}
    \caption{Experimental Protocols for the slip detection of the mobile robot}
    \label{fig:protocol}
\end{figure}

We designed the experiment such that sensor data of situations where a mobile robot drops an object while moving can be easily reproduced and recorded.
The common protocol is depicted in Fig. \ref{fig:protocol}.
First, the robot extends its arm and gripper, and grasps the object that the experimenter delivers.
3 seconds after holding the object, the robot starts recording data and moves in a predefined pattern.
After 5 seconds, the robot releases the force of the gripper while moving in order to drop the object.
The robot stops recording the sensor data 500 ms after the gripper is opened.
This extra 500 ms is a sufficient amount of time for the dropped object to hit the ground.
The data collected before the drop is labeled as normal, and the data collected
in the 500 ms after dropping the object is labeled as abnormal.

Robots can drop objects for a variety of reasons including bumping into an obstacle or swaying due to uneven flooring.
However, for controlled analysis, we only consider the sensor output during an object slip situation, regardless of the cause.

\begin{figure}[t]
    \centering
    \includegraphics[width=1.0\columnwidth]{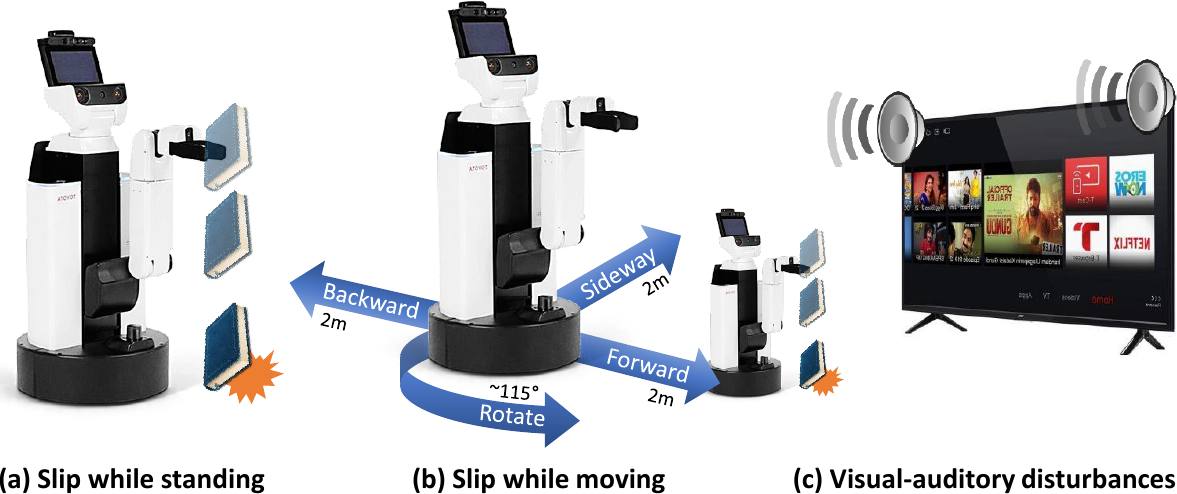}
    \caption{Three experimental configurations according to different noise types}
    \label{fig:noises}
\end{figure}

The experiment was conducted on 8 objects in Table \ref{tab:datasets}, with 4 moving patterns depicted in Fig. \ref{fig:protocol} and \ref{fig:noises}(b).
The moving patterns included moving forward, backward, sideways, and in-place rotation.
We used these moving patterns so that the model could be trained for general moving scenarios and not just monotonous movements.
Experiments for all possible object and moving pattern combinations were repeated 30 times.
This resulted in approximately 56,000 sets of sensor data being collected from a total of 960 trials.
We divided this into a training set (30.8K), validation set (10.2K), and evaluation set (15K).
The starting position of the robot and the position and orientation of the object being held were purposely altered each time, so as to ensure the model could learn about general situations.

An important part of the experiment was to mimic noisy real-world environments effectively.
The irregular movement of the experimenters and coworkers walking around and talking during the experiment provided natural noise to the sensors.
However, this noise was not enough to truly simulate the noise of real-world public environments.
Therefore, in order to create additional visual and auditory noise, we placed a 40-inch monitor as in Fig. \ref{fig:noises}(c) and played videos randomly chosen by Youtube which are not static, such as sportscasting or music video.

In order to build a set of consistent input data, sensor data from unsuccessful trials were discarded and the experiments were repeated.
For example, the furry toy's ears were caught in the gripper and did not fall even after the gripper was opened.
In other trials, the robot failed to execute its target movements as expected due to errors on localization.

\subsection{Experimental Configuration According to Noise}
We conduct ablation study to examine that how the noise from robot movement and visual-auditory disturbances affects the anomaly detection performance on each sensor data.
To this end, we compare slip detection performance of three experimental configurations according to different noise source combination, as illustrated in Fig.$\,$\ref{fig:noises}.

First, in the standing condition, the robot performed the object slip task without moving or the presence of visual-auditory disturbances as in Fig. \ref{fig:noises}(a).
Secondly, in the moving condition, the robot performed the task while moving in the same pattern as in Fig. \ref{fig:noises}(b), but visual and auditory noises were not given.
Lastly, in the visual-auditory disturbed condition, the robot performed the task while moving with visual-auditory disturbance from random video.

\section{Experimental Results}

\subsection{Performance Comparison between Multimodal and Unimodal Approaches}
\begin{figure}[H]
\centering
    
    \includegraphics[width=.92\columnwidth]{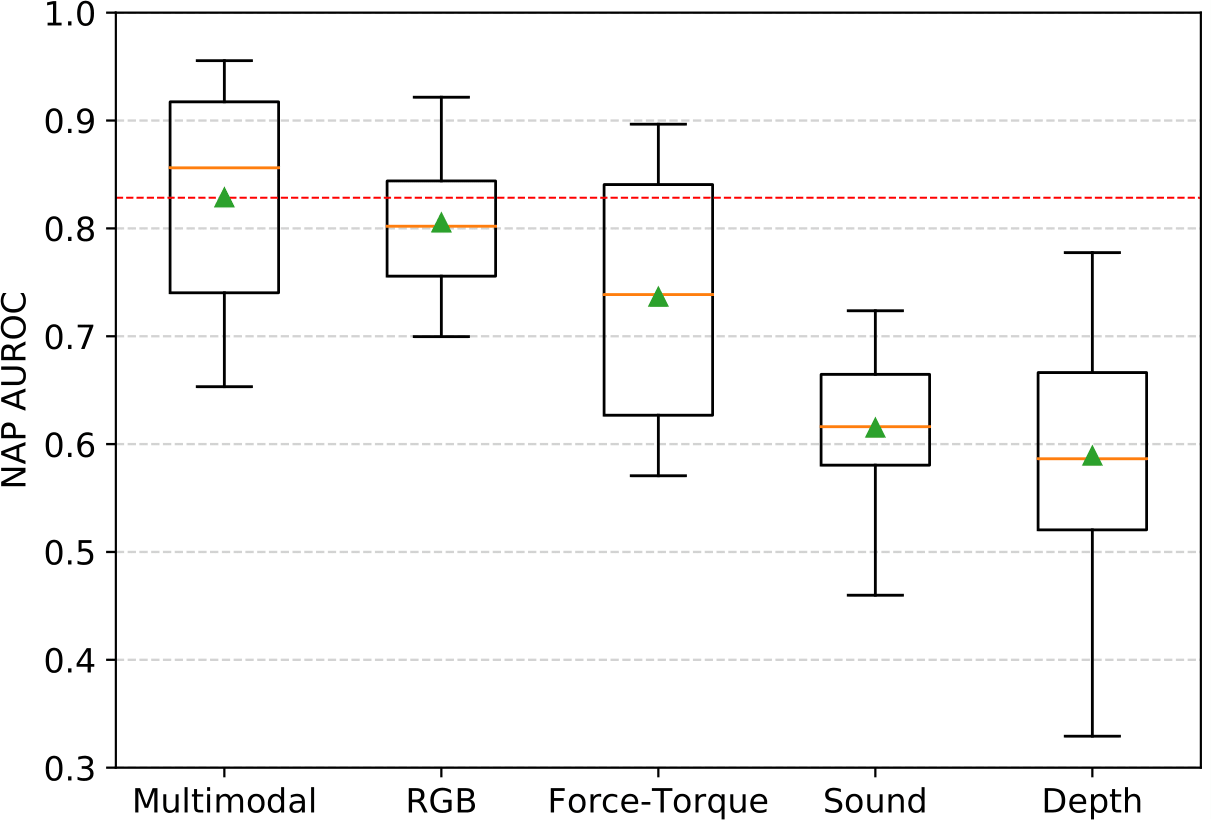}
    \vspace{-.1cm}
    \caption{Anomaly detection results by sensor type. Orange lines and green triangles in each box indicate the median and mean values, respectively, and the red dashed line stands for the mean value of multimodal-based results.}
    \label{fig:result_nap}
    
\vspace{.5cm}

    \includegraphics[width=.93\columnwidth]{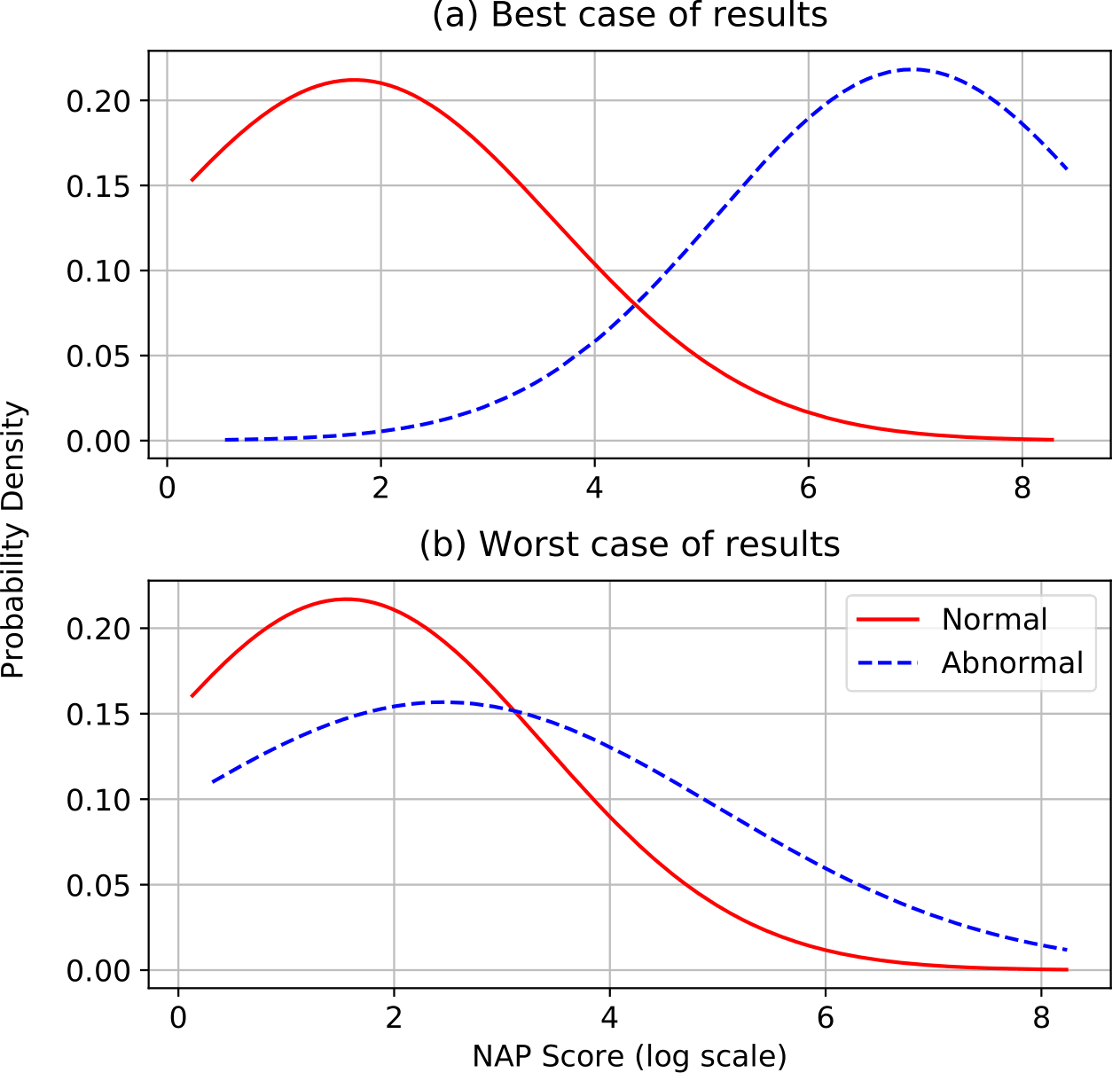}
    \vspace{-.1cm}
    \caption{Comparison of distributions of NAP score between normal and abnormal samples in the best or the worst cases.}
    \label{fig:result_dist}

\vspace{.5cm}

    \includegraphics[width=.93\columnwidth]{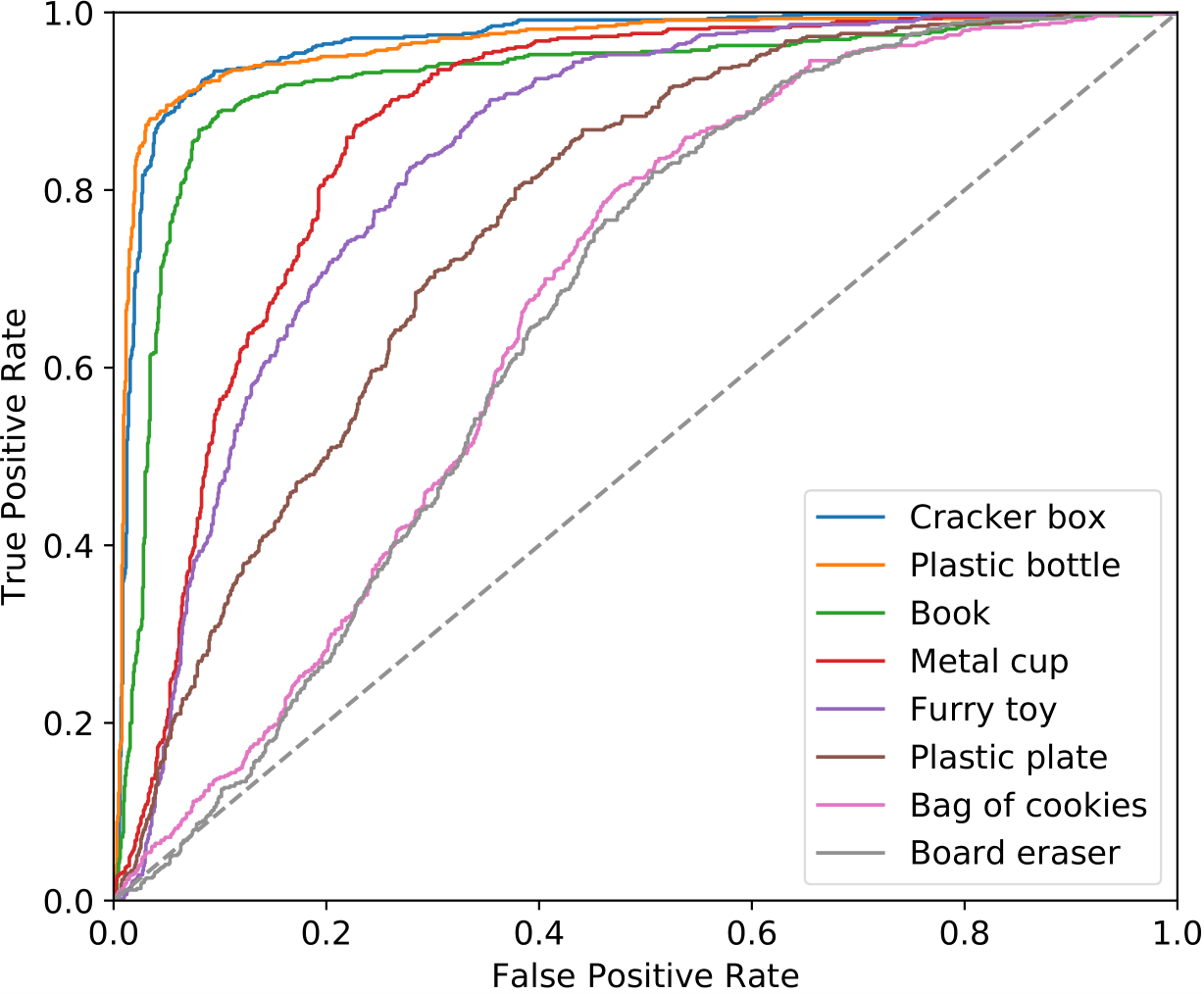}
    \vspace{-.1cm}
    \caption{ROC curves of NAP scores by object type. The closer the graph is to the upper left corner, the higher the performance.}
    \label{fig:result_roc}
\end{figure}

Results comparing the performance of anomaly detection using multimodal and unimodal data are shown in Fig. \ref{fig:result_nap}.
The larger NAP AUROC value indicates that the abnormal data is more clearly distinguishable from the normal data.
NAP AUROC scores computed by multimodal data outperformed scores of any other unimodal data with a mean score of 0.8329.
In unimodal data, the NAP AUROC scores of using the RGB and force-torque sensor showed relatively high performance, and the sound and depth showed lower scores.

The results can also be confirmed intuitively by looking at the distributions of the NAP score of multimodal data as shown in Fig. \ref{fig:result_dist}.
In the best case which showed the highest NAP AUROC, the anomaly scores of the normal sample were distributed at a low value and the scores of the abnormal sample were distributed at a high value, showing that the two distributions were clearly separated.
However, in the worst case with the lowest NAP AUROC, the two distributions almost overlap as the anomaly scores of both the normal and abnormal samples are distributed at low values.

Analysis of the results for each object confirms that the performance of the model varied for each type of object as shown in Fig. \ref{fig:result_roc}.
Overall, positive results were shown for all the objects (over 0.65), while the lighter weight objects such as board eraser, bag of cookies, and plastic plate showed low performance.
It is worth noting, the force-torque sensor showed a lower NAP AUROC value than that of RGB in Fig. \ref{fig:result_nap}, but a higher NAP AUROC value for heavier objects.
This reveals that the force-torque sensor played a significant role in detecting object slip, as the drop of heavier objects were better detected.


\subsection{Performance Comparison According to Different Noise Configurations}
The performance of the proposed method according to the various noise settings are shown in Table \ref{table:comparison-noises}.
In this experiment, in order to focus on the effect of noise on each sensor, only the results for one object, the book is presented.
The book was selected after considering the size, weight, and anomaly scores of each object.
Note that the 90\% quantile of the NAP score of each validation set was used for the threshold of anomaly detection to calculate the F1 score.

\begin{table*}[!htb]
    \caption{Comparison of model performance by multimodal and unimodal data under different experimental conditions and metrics}
    \centering
    \renewcommand\arraystretch{1.1}
    \resizebox{\textwidth}{!}{
        \tiny 
        \begin{tabular}{@{}cccccccccc@{}}
            \toprule
            \multirow{2}{*}{Sensors}
          & \multicolumn{3}{c}{AUROC}
          & \multicolumn{3}{c}{AUPRC}
          & \multicolumn{3}{c}{F1 Score} \\
            \cmidrule(l{4pt}r{4pt}){2-4}
            \cmidrule(l{4pt}r{4pt}){5-7}
            \cmidrule(l{4pt}r{0pt}){8-10}
          & Standing & Moving & V.A.D.$^\star$ & Standing & Moving & V.A.D. & Standing & Moving & V.A.D. \\
            \midrule
            Multimodal & \textbf{0.9904} & \textbf{0.9323} & \textbf{0.9199} & \textbf{0.9883} & \textbf{0.8276} & \textbf{0.7865} & \textbf{0.8940} & \textbf{0.8188} & \textbf{0.8342}  \\
            Force-Torque & 0.9867 & 0.6589 & 0.6681 & 0.9832 & 0.4006 & 0.4107 & 0.8891 & 0.2032 & 0.2173  \\
            RGB & 0.9580 & 0.8762 & 0.7826 & 0.9096 & 0.7236 & 0.6616 & 0.8729 & 0.6559 & 0.5826  \\
            Depth & 0.9309 & 0.8456 & 0.5207 & 0.9105 & 0.7747 & 0.3565 & 0.7747 & 0.7571 & 0.2227  \\
            MIC & 0.9188 & 0.7264 & 0.6490 & 0.8884 & 0.6508 & 0.4884 & 0.7970 & 0.5486 & 0.3662  \\
            \bottomrule
            \multicolumn{10}{l}{\scalebox{.8}{$^\star$ V.A.D. stands for visual-auditory disturbance}}
        \end{tabular}
    }
    \label{table:comparison-noises}
\end{table*}

\begin{figure*}[!htb]
\centering
\includegraphics[width=\linewidth]{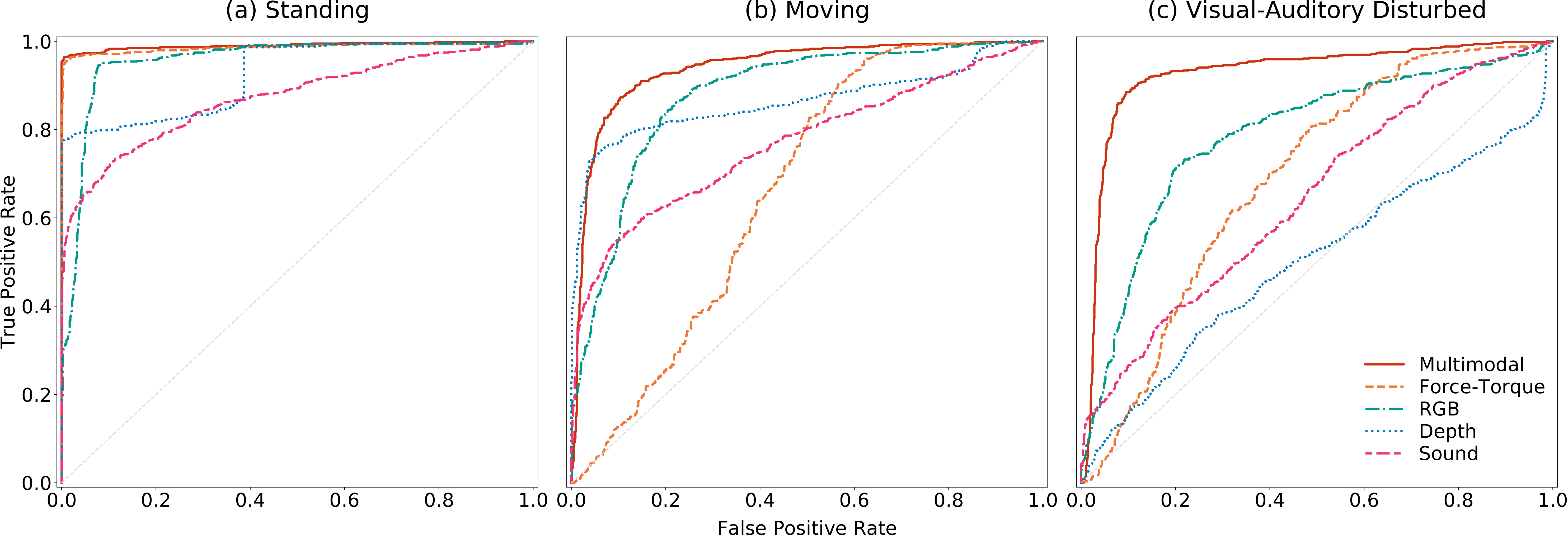}
\vspace{-.5cm}
\caption{Comparison of ROC curves under different experimental conditions. Multimodal data intuitively shows the most noise-resistant performance.}
\label{fig:result_roc_sensor_wised}
\end{figure*}


It was found that the overall anomaly detection performance decreased as motion and visual-auditory noise was added for all sensors.
However, the performance of multimodal data showed a very small decrease compared to other sensors, and proved that it is robust to environmental noise.

Performance differences between the sensors can be seen prominently through the NAP ROC curve in Fig. \ref{fig:result_roc_sensor_wised}.
Overall the performance decreased with the addition of noise, especially with the visual-auditory disturbance.
The performance on the force-torque sensor data declined significantly in the moving conditions compared to other sensors, but was not affected in the visual-auditory disturbance.
It is assumed that the stable performance on multimodal data is due to different sensors complementarily providing the cues required for anomaly detection.
This compensates for the performance degradation of other sensors affected by the diverse environmental noise.


%
%
\section{Conclusion}

In this paper, we proposed an autoencoder-based multimodal anomaly detection framework for robust slip perception in mobile manipulation robots.

To use heterogeneous sensor data of different sizes and properties as the multimodal input of the autoencoder, we added an integration process that performs synchronization, normalization and integration of data from all sensors in the proposed framework.

We verified the proposed method with the multisensory data collected by experiments using a mobile robot grasping and dropping diverse household objects while moving in a real-world environment.
Visual and auditory noise from video and moving people were randomly interrupted each sensor of the robot during the experiments.

Our experimental results showed that the proposed method performed noticeably better with multimodal sensors compared to using single sensors independently.
From ablation studies according to different noise configurations, we assumed that each sensor data can complement each other's noise.

The consumed time to perform the proposed method is a total of 29.3 ms: 12.0 ms for the integration of multisensory data, 2.7 ms for the autoencoder hidden layer restoration value generation, and 14.6 ms for the NAP calculation. So, it is confirmed that our method can be used in real-time.

However, there still exists room for extended research, as our experimental setup was relatively static compared to actual public areas, even with the application of artificial noise settings.
Furthermore, a more diverse pool of objects will have to be evaluated in order to assess the generalizability of the proposed model.

\bibliographystyle{IEEEtran}
\bibliography{root}

\end{document}